\pdfoutput=1

\documentclass{article}

\usepackage[main, preprint]{neurips_2026}

\usepackage[utf8]{inputenc} 
\usepackage[T1]{fontenc}    
\usepackage{hyperref}       
\usepackage{url}            
\usepackage{booktabs}       
\usepackage{amsmath}        
\usepackage{amsfonts}       
\usepackage{nicefrac}       
\usepackage{microtype}      
\usepackage{xcolor}         
\usepackage{tabularx}
\usepackage{caption}
\usepackage{subcaption}
\usepackage{orcidlink}
\usepackage{algorithm}
\usepackage{algpseudocode}

\begin{document}

\title{Language steering in latent space to mitigate unintended code-switching}

\author{
  Andrey Goncharov\orcidlink{0009-0006-9471-2181}\\
  Applied AI Institute\\
  \texttt{andrey@faillearnrepeat.net}
  \And
  Nikolai Kondusov\\
  Voronezh State University
  \And
  Alexey Zaytsev\orcidlink{0000-0002-1653-0204}\\
  Applied AI Institute
}

\maketitle

\begin{abstract}
  Multilingual Large Language Models (LLMs) often exhibit hallucinations such as unintended code-switching, reducing reliability in downstream tasks.
  We propose latent-space language steering, a lightweight inference-time method that identifies a language direction via Principal Component Analysis (PCA) on parallel translations
  and steers embeddings along this axis to control language identity.
Our approach reduces next-token distributional divergence by up to 55\%, Code-Switching Index by 63--99\%, while preserving broad semantic direction with negligible computational overhead
and requires only minimal parallel data for calibration.
\end{abstract}


\section{Introduction}

\begin{figure}[ht]
  \centering
  \includegraphics[width=1\linewidth]{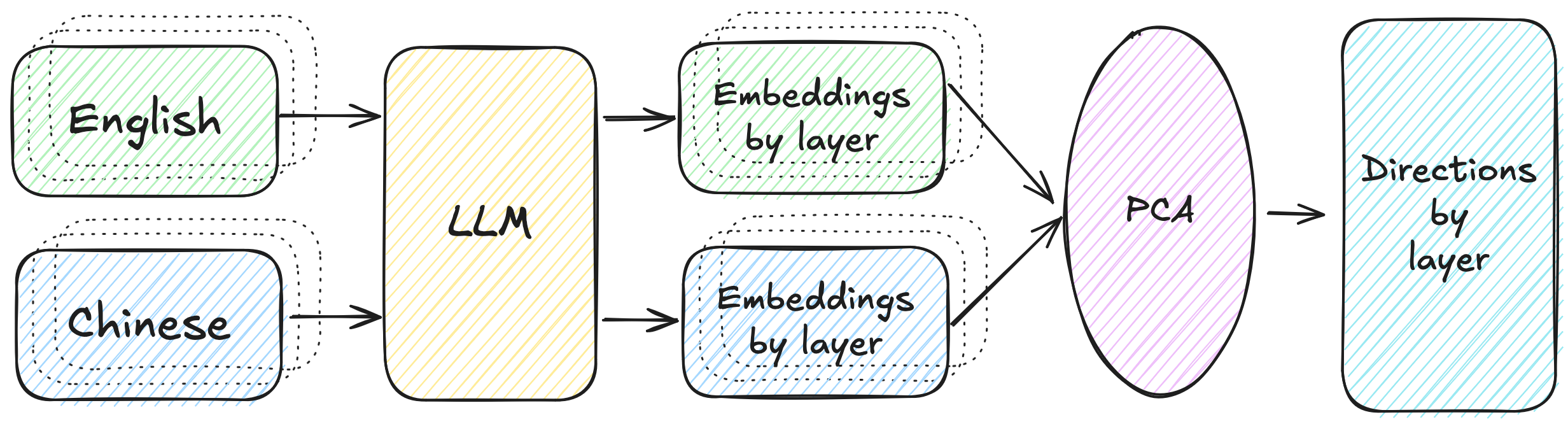}
  \caption{Language steering in hidden space: 1. Collect parallel translations of the same concepts; 2. LLM forward pass to get embeddings by layer; 3. Perform PCA by layer on the merged embeddings.}
  \label{fig:lang_steering_scheme}
\end{figure}

LLMs exhibit unintended code-switching - generating tokens in languages other than the target~\citep{marchisio-etal-2024-understanding}.
\citet{ryan-etal-2024-unintended} document how LLM alignment procedures can cause models to default to English even when prompted in other languages,
\citet{yoo2025codeswitchingredteamingllmevaluation} demonstrate that code-switching serves as an effective red-teaming vector, bypassing safety filters.
Existing solutions require costly fine-tuning~\citep{marchisio-etal-2024-understanding} or introduce latency via decoding-time interventions~\citep{dathathri2020plugplaylanguagemodels, yang-klein-2021-fudge, liu2021dexpertsdecodingtimecontrolledtext}.

We propose \emph{latent space language steering}, a cheap inference-time method that exploits low-dimensional language directions in hidden representations.
By applying PCA to parallel translations, we identify linear subspaces separating languages while preserving broad semantic direction.
Steering via simple projection (one dot product and vector subtraction per token) mitigates code-switching with negligible overhead.

Additionally, we analyze language clustering dynamics across layers and discover that a single final layer projection enables near-perfect language classification. Code and data are released to facilitate further research \footnote{\url{https://github.com/fxlrnrpt/language-steering-in-latent-space}}.

\section{Related works}

\label{sec:related}

\paragraph{Code-switching and multilingual control.}
While code-switching has been studied for mixed-language inputs in speech and NER \citep{aguilar-etal-2020-lince, bali-etal-2014-borrowing} and in LLMs \citep{zhang-etal-2023-multilingual},
recent work identifies unintended code-switching in multilingual LLM outputs \citep{ryan-etal-2024-unintended, yoo2025codeswitchingredteamingllmevaluation},
degrading user experience and evaluation.
Standard mitigations require costly fine-tuning on monolingual data \citep{marchisio-etal-2024-understanding}, reinforcement learning with language-specific rewards \citep{Guo_2025}, or brittle prompt engineering \citep{marchisio-etal-2024-understanding}.

\paragraph{Language structure in representations.}
Work on mBERT and XLM-R showed language identity concentrates in few principal components, separable from semantics in parallel texts \citep{conneau-etal-2020-emerging, chi-etal-2020-finding, libovicky-etal-2020-language}.
\citet{yang-etal-2021-simple} removed these components for language-agnostic embeddings.
\citet{wendler-etal-2024-llamas} extended this to decoder-only LLMs, revealing latent language preferences, while language-specific neurons concentrate in the first and last layers \citep{tang-etal-2024-language, kojima-etal-2024-multilingual, zhao2024largelanguagemodelshandle}.
Language-specific subspaces \citep{zhao2025lensrethinkingmultilingualenhancement} and sparse autoencoder features \citep{deng-etal-2025-unveiling} have also been used to enhance or steer multilingual generation.
We exploit this structure for active generation control.

\paragraph{Representation steering and concept erasure.}
Recent work has shown that high-level concepts can be located and manipulated in model representations via linear methods \citep{pmlr-v235-park24c}.
\citet{zou2025representationengineeringtopdownapproach} introduced representation engineering, using PCA on contrastive prompt pairs to identify concept directions (e.g., honesty, harmlessness) and steering activations along them.
\citet{turner2024steeringlanguagemodelsactivation} proposed activation addition, adding steering vectors to the residual stream at inference time, later extended to contrastive activation addition \citep{rimsky-etal-2024-steering}.
In parallel, concept erasure methods like \citet{ravfogel-etal-2020-null} (iterative null-space projection) and ~\citet{belrose2025leaceperfectlinearconcept} (closed-form linear erasure) remove specific attributes from representations by projecting onto their null space.
\citet{li2024inferencetimeinterventionelicitingtruthful} applied probing-guided inference-time intervention on attention heads to elicit truthful outputs.
These methods have primarily targeted sentiment, truthfulness, and toxicity.
We apply the same conceptual framework to language identity.

\section{Methods}

Our approach operates in the latent space of pre-trained multilingual LLMs without fine-tuning.
We identify language-specific directions via PCA on parallel translations, then steer representations along these directions to control code-switching (Algorithm~\ref{alg:steering}).

\subsection{Language Direction Identification}

Given a parallel corpus $\mathcal{D} = \{(s_i, k_i)\}_{i=1}^N$ where $s_i$ represents semantically equivalent content in language $k_i \in \mathcal{K}$,
we extract hidden states $\mathbf{h}_i^{(\ell)} \in \mathbb{R}^d$ from each layer $\ell$ of the LLM.
Specifically, we use the hidden states of all tokens of each sentence, excluding special tokens and the first token.
For each layer, we pool hidden states across all languages and apply PCA to the centered embeddings $\mathcal{H}^{(\ell)} = \{\mathbf{h}_i^{(\ell)}\}_{i=1}^M$, where $M$ is the total number of tokens.
The first principal component $\mathbf{v}^{(\ell)}$ captures the language direction:
\begin{equation}
  \mathbf{v}^{(\ell)} = \arg\max_{\|\mathbf{v}\|=1} \sum_{i=1}^M \left(\mathbf{v}^\top (\mathbf{h}_i^{(\ell)} - \bar{\mathbf{h}}^{(\ell)})\right)^2
\end{equation}
When semantic content is constant across languages, language identity becomes the dominant variance source, making $\mathbf{v}^{(\ell)}$ an interpretable language axis.

\subsection{Latent-Space Language Steering}

Empirical analysis (Section~\ref{sec:results}) shows language variance concentrates in final layers, consistent with prior work \citep{tang-etal-2024-language, zhao2024largelanguagemodelshandle}.
We apply steering only to layers $\ell \geq \ell_{\text{crit}}$: for Qwen2.5-1.5B (28 layers), we steer layers 27-28; for Llama-3.2-1B (16 layers), we steer layers 15-16.
For each token's hidden state $\mathbf{h}_t^{(\ell)}$ during generation, we modify the language component:
\begin{equation}
  \tilde{\mathbf{h}}_t^{(\ell)} = \mathbf{h}_t^{(\ell)} + s \left((\mathbf{h}_t^{(\ell)} - \bar{\mathbf{h}}^{(\ell)}) \cdot \mathbf{v}^{(\ell)}\right) \mathbf{v}^{(\ell)}
\end{equation}
where $s \in \mathbb{R}$ controls intervention strength.

\begin{algorithm}[t]
  \caption{Latent-space language steering}
  \label{alg:steering}
  \begin{algorithmic}[1]
    \Require LLM, parallel corpus $\mathcal{D}$, steered layers $\{\ell : \ell \geq \ell_{\text{crit}}\}$, coefficient $s$
    \Statex \textit{Calibration (once per language pair):}
    \For{each steered layer $\ell$}
    \State $\mathcal{H}^{(\ell)} \gets$ hidden states $\mathbf{h}_i^{(\ell)}$ of all tokens of all $s_i \in \mathcal{D}$ \Comment{excl.\ special and first tokens}
    \State $\bar{\mathbf{h}}^{(\ell)} \gets \frac{1}{M} \sum_{i=1}^{M} \mathbf{h}_i^{(\ell)}$
    \State $\mathbf{v}^{(\ell)} \gets$ first principal component of $\{\mathbf{h}_i^{(\ell)} - \bar{\mathbf{h}}^{(\ell)}\}_{i=1}^{M}$
    \EndFor
    \Statex \textit{Inference (every forward pass):}
    \For{each steered layer $\ell$, after the layer computes $\mathbf{h}_t^{(\ell)}$ for every new token $t$}
    \State $p_t \gets (\mathbf{h}_t^{(\ell)} - \bar{\mathbf{h}}^{(\ell)}) \cdot \mathbf{v}^{(\ell)}$ \Comment{Project to language direction}
    \State $\mathbf{h}_t^{(\ell)} \gets \mathbf{h}_t^{(\ell)} + s \, p_t \, \mathbf{v}^{(\ell)}$
    \EndFor
  \end{algorithmic}
\end{algorithm}

\section{Results}
\label{sec:results}

\begin{figure*}[ht]
  \centering
  \begin{subfigure}[b]{0.48\linewidth}
    \centering
    \includegraphics[width=1\linewidth]{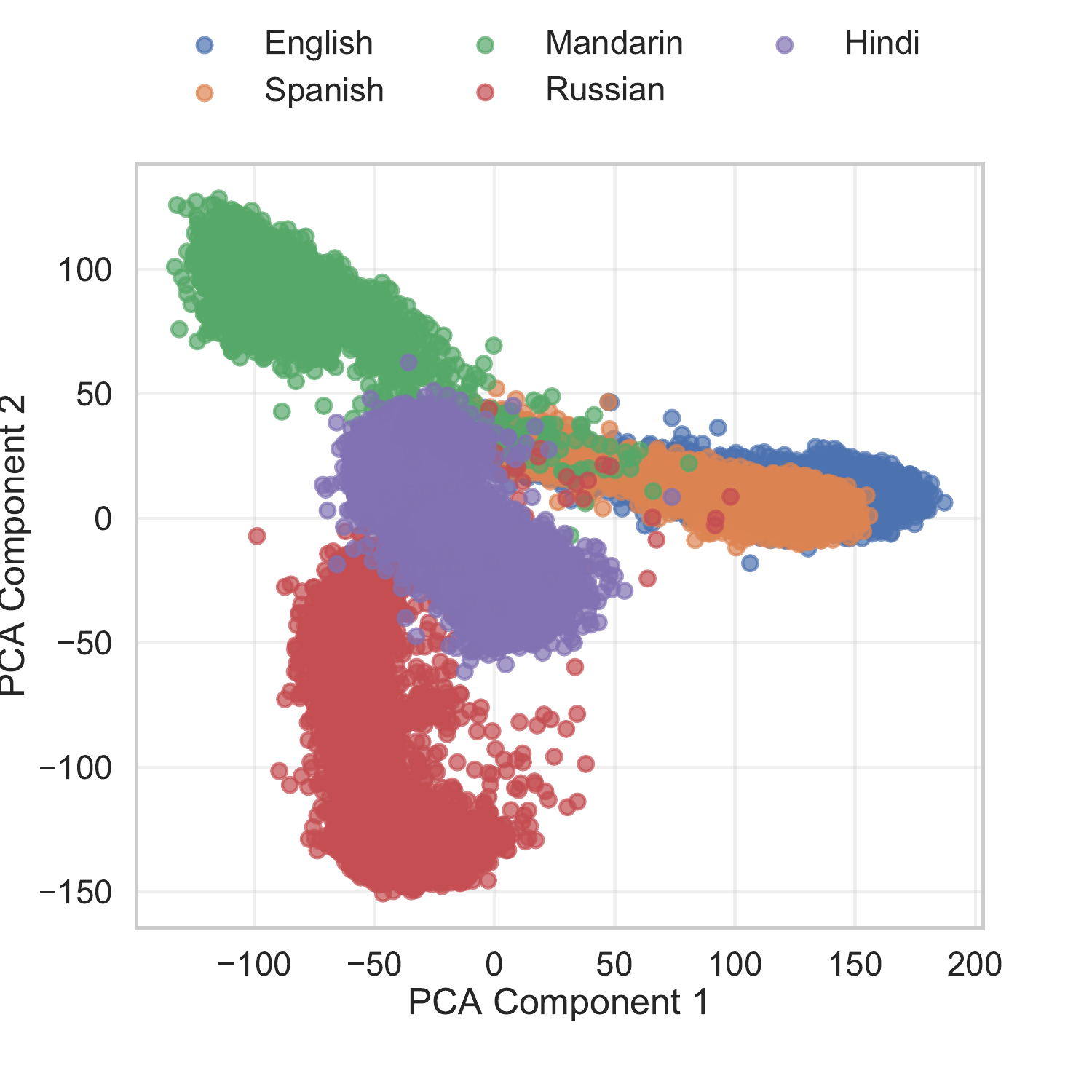}
    \caption{Qwen}
    \label{fig:qwen_lang_map}
  \end{subfigure}
  \hfill
  \begin{subfigure}[b]{0.48\linewidth}
    \centering
    \includegraphics[width=1\linewidth]{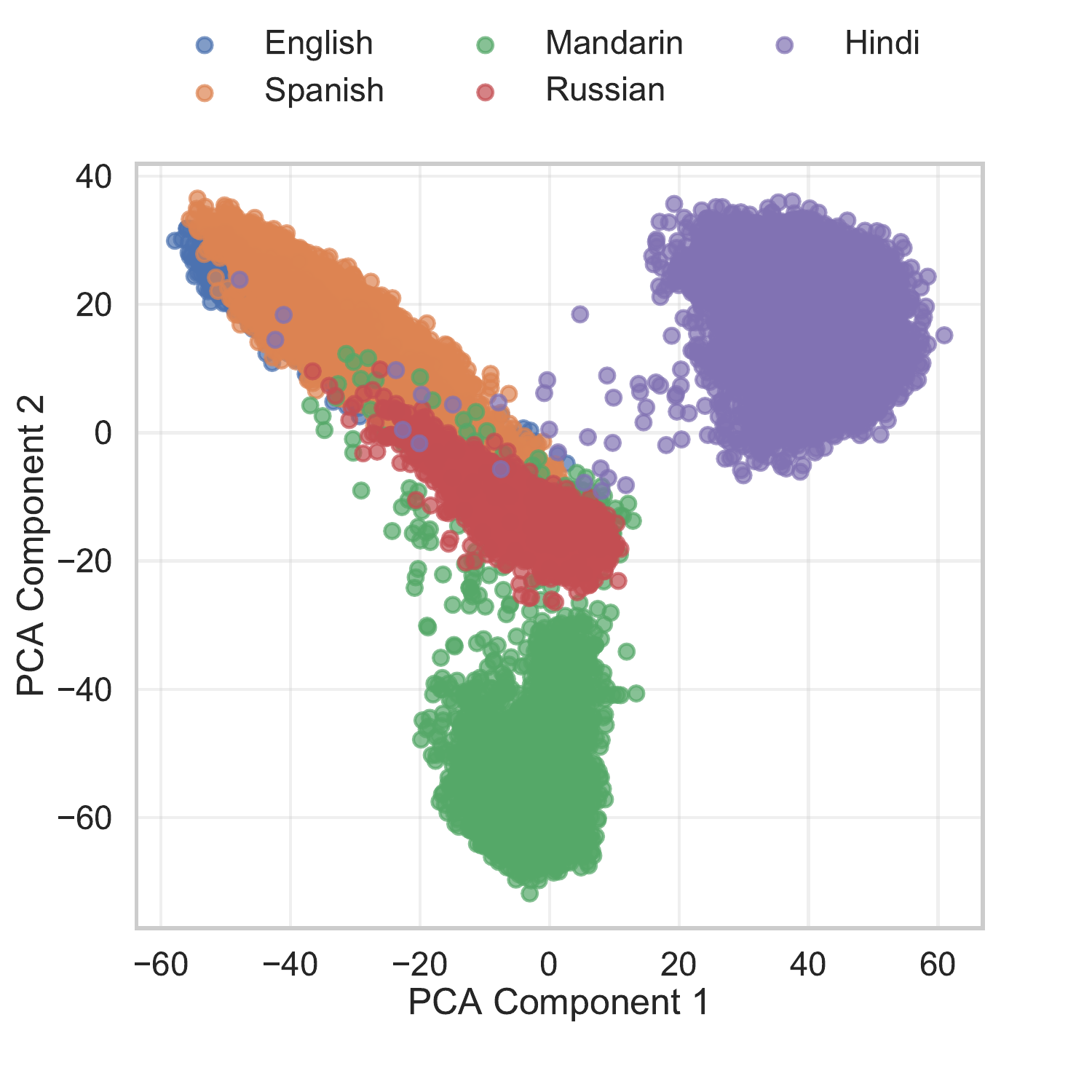}
    \caption{Llama}
    \label{fig:llama_lang_map}
  \end{subfigure}
  \caption{Final layer language map}
\end{figure*}

\begin{figure*}[ht]
  \centering
  \begin{subfigure}[t]{0.24\linewidth}
    \centering
    \includegraphics[width=1\linewidth]{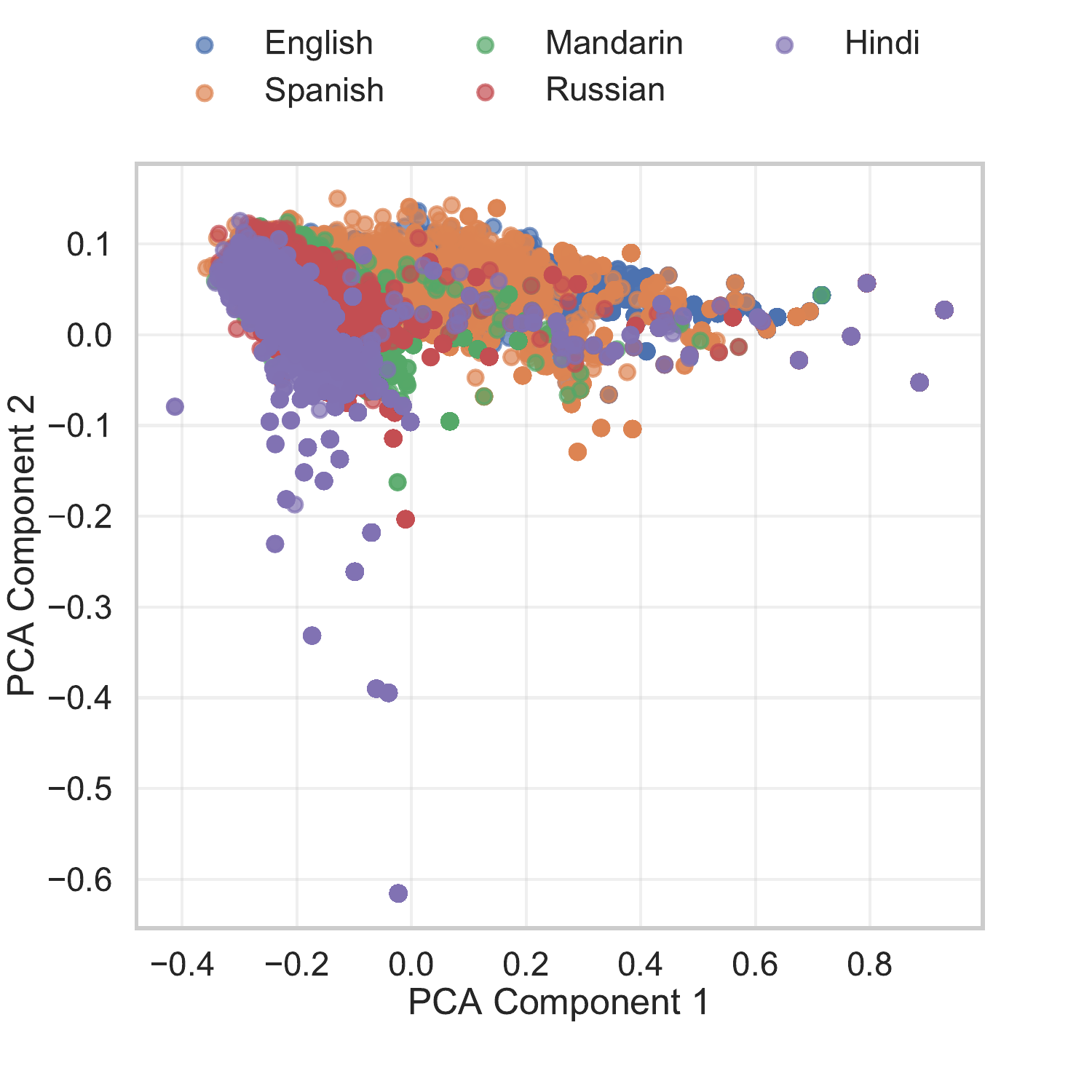}
    \caption{First layer}
    \label{fig:llama_lang_map_first}
  \end{subfigure}
  \hfill
  \begin{subfigure}[t]{0.24\linewidth}
    \centering
    \includegraphics[width=1\linewidth]{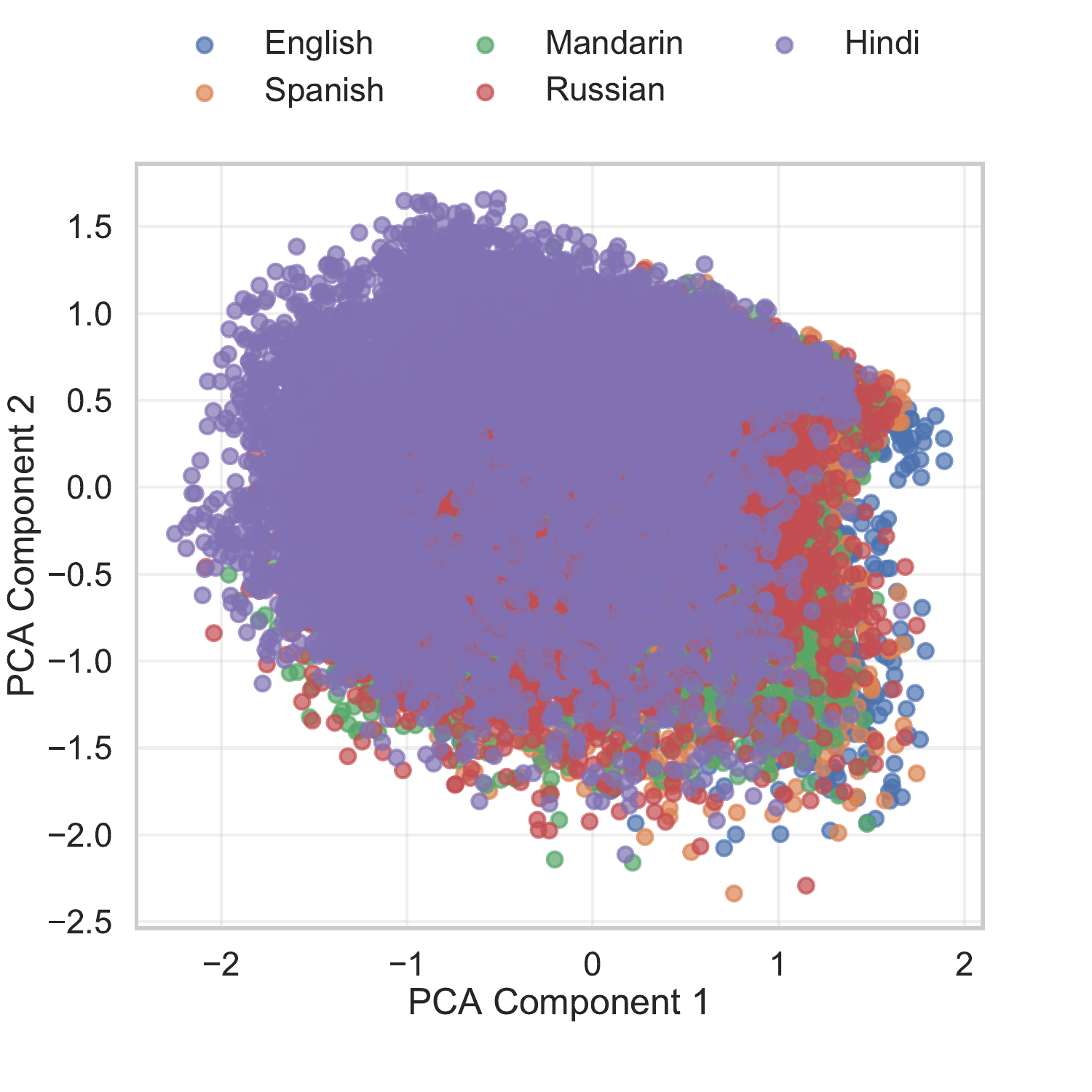}
    \caption{Mid layer}
    \label{fig:llama_lang_map_mid}
  \end{subfigure}
  \hfill
  \begin{subfigure}[t]{0.48\linewidth}
    \centering
    \includegraphics[width=1\linewidth]{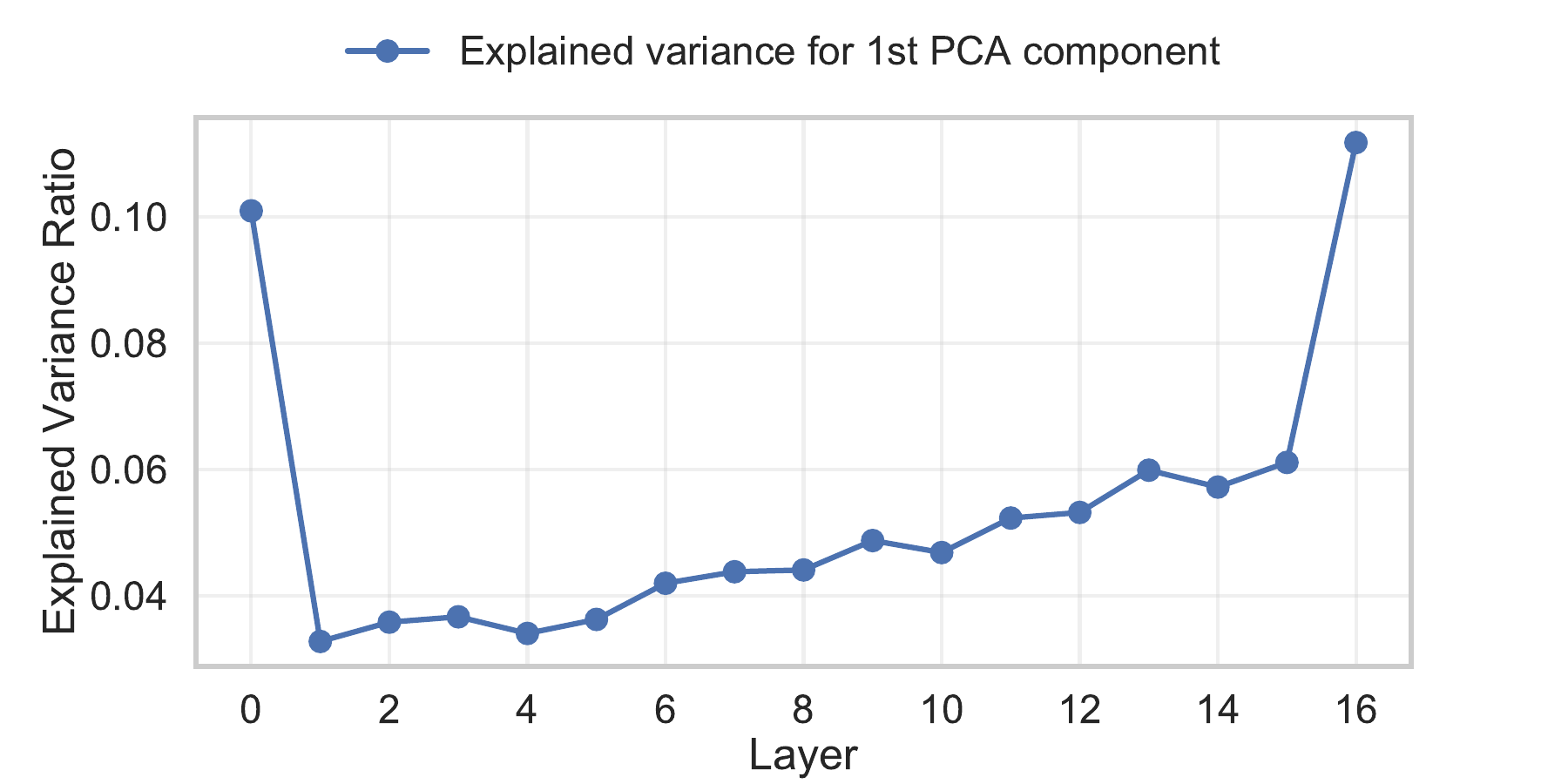}
    \caption{Explained variance}
    \label{fig:llama_explained_variance}
  \end{subfigure}
  \caption{Llama layer-wise language maps and explained variance}
\end{figure*}

\begin{figure*}[ht]
  \centering
  \begin{subfigure}[t]{0.24\linewidth}
    \centering
    \includegraphics[width=1\linewidth]{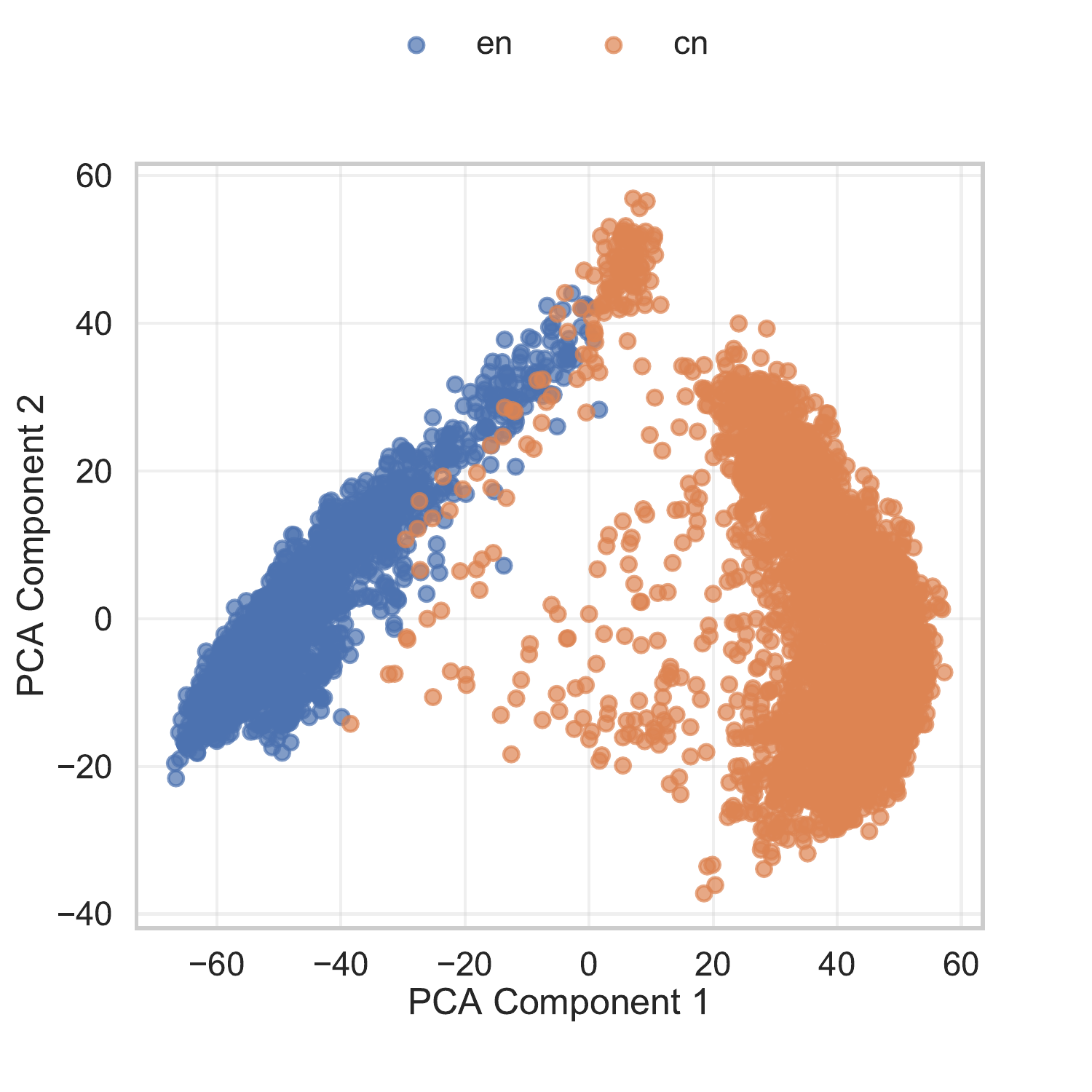}
    \caption{English - Chinese}
    \label{fig:llama_lang_map_en_cn}
  \end{subfigure}
  \hfill
  \begin{subfigure}[t]{0.24\linewidth}
    \centering
    \includegraphics[width=1\linewidth]{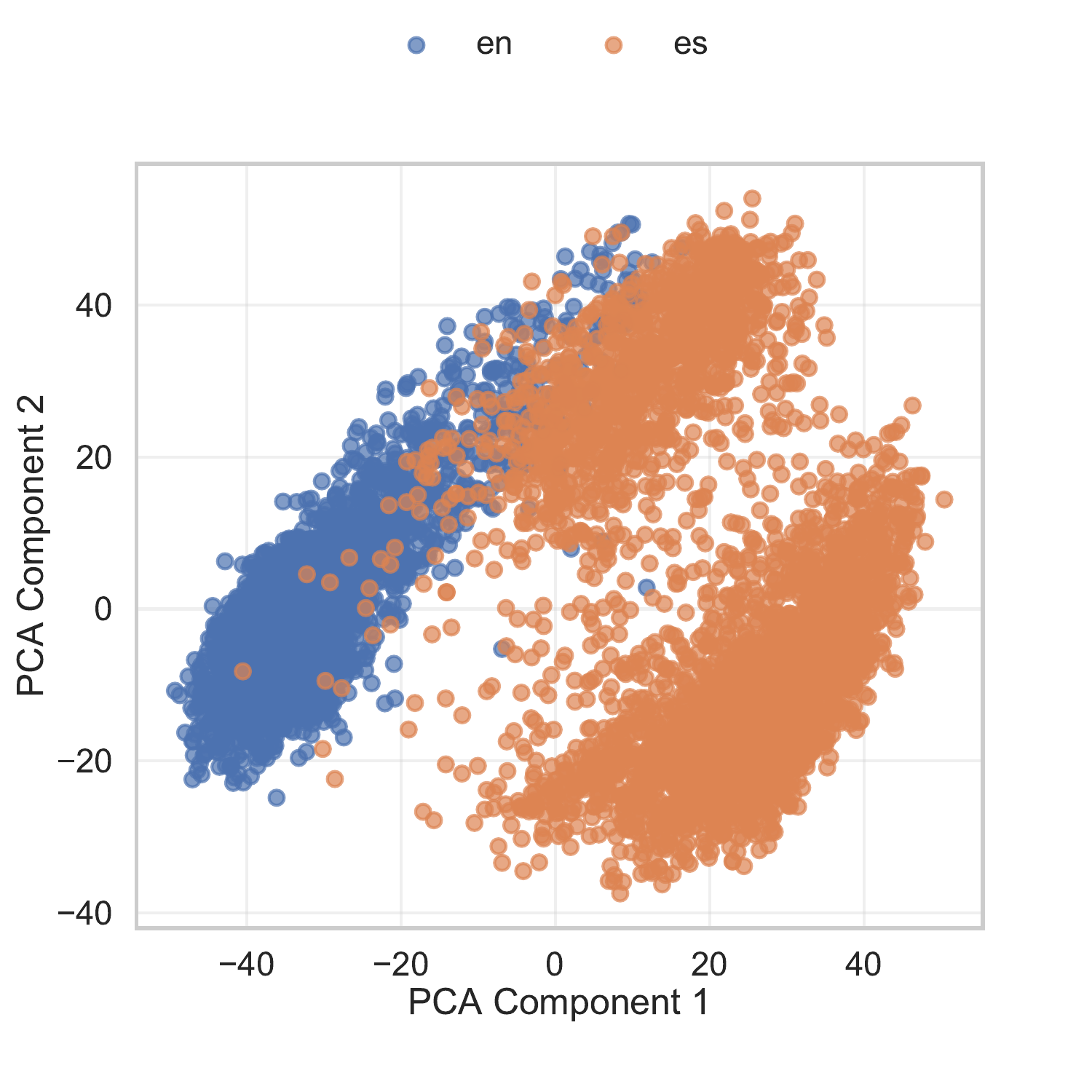}
    \caption{English - Spanish}
    \label{fig:llama_lang_map_en_es}
  \end{subfigure}
  \hfill
  \begin{subfigure}[t]{0.24\linewidth}
    \centering
    \includegraphics[width=1\linewidth]{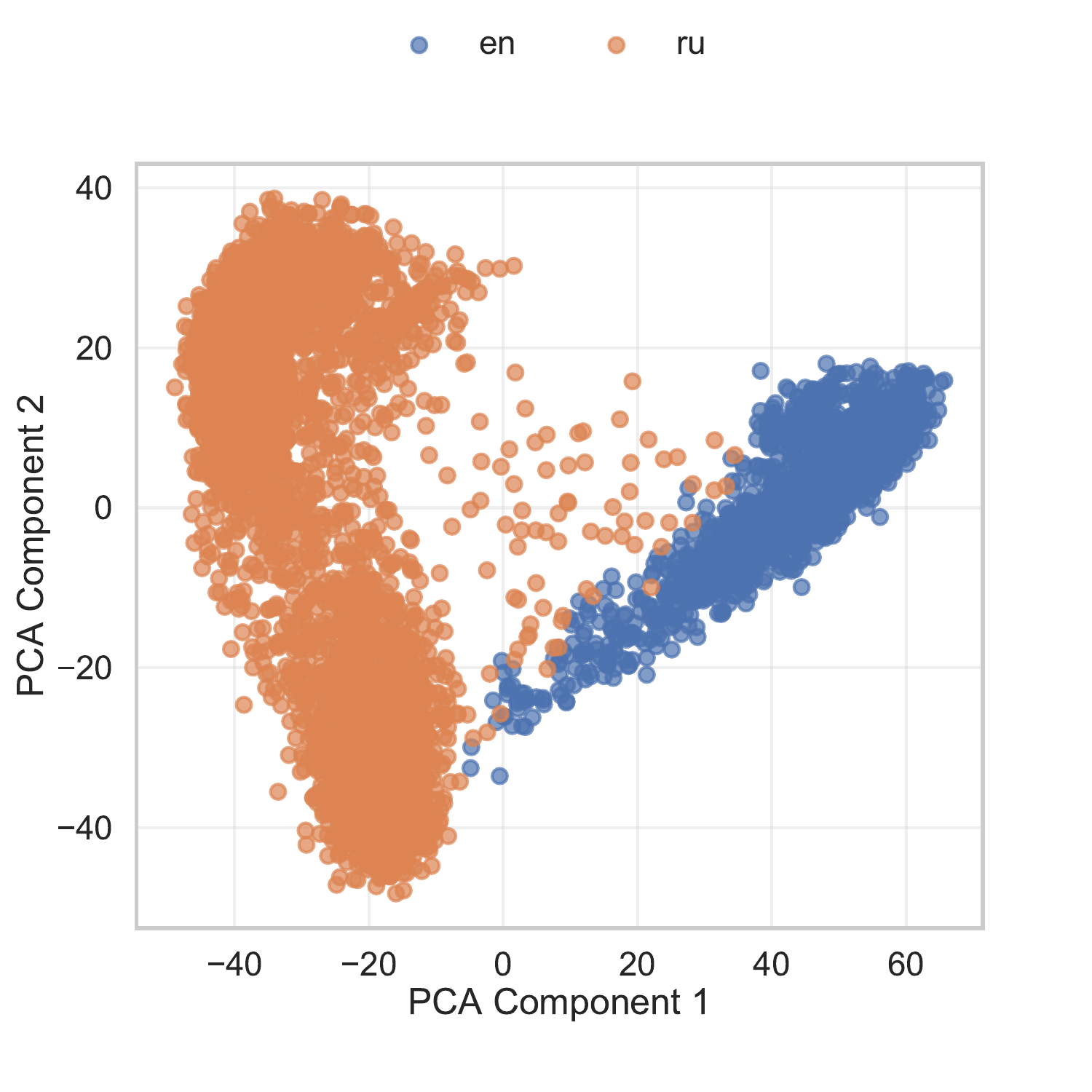}
    \caption{English - Russian}
    \label{fig:llama_lang_map_en_ru}
  \end{subfigure}
  \hfill
  \begin{subfigure}[t]{0.24\linewidth}
    \centering
    \includegraphics[width=1\linewidth]{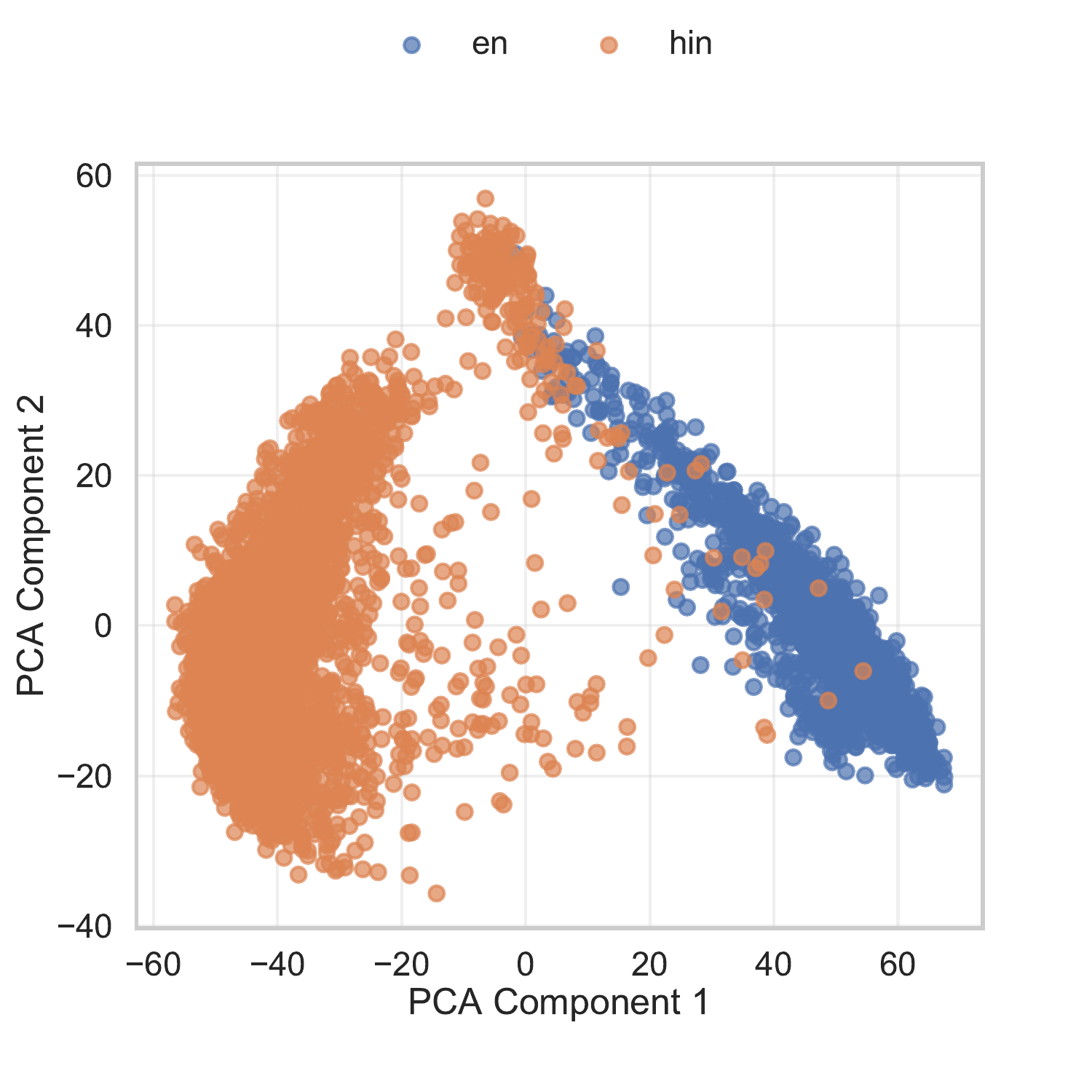}
    \caption{English - Hindi}
    \label{fig:llama_lang_map_en_hin}
  \end{subfigure}
  \caption{Llama final layer language maps for each language pair}
\end{figure*}

\subsection{Language direction identification}

\paragraph{Experimental setup.} Dataset: Flores Plus \cite{scaling-2024} with 50 parallel samples (English, Spanish, Russian, Chinese, Hindi) for PCA fitting and 100 for validation. Models: Qwen2.5-1.5B ~\cite{qwen2025qwen25technicalreport} and Llama-3.2-1B ~\cite{grattafiori2024llama3herdmodels}.

\paragraph{Language maps reveal emergent clustering.} Figure~\ref{fig:qwen_lang_map} and~\ref{fig:llama_lang_map} show final-layer embeddings projected onto the first two PCs.
Consistent with prior findings~\cite{conneau-etal-2020-emerging, wendler-etal-2024-llamas}, models organize identical content into tight, language-specific clusters with clear boundaries.
Languages within the Indo-European family (Spanish, Russian) cluster closer than typologically distant pairs (English--Chinese), suggesting PC1 captures coarse language distinction while PC2 encodes finer linguistic properties.

Figures ~\ref{fig:llama_lang_map_first} ~\ref{fig:llama_lang_map_mid} show layer-wise evolution:
clusters emerge loosely in early layers and sharpen dramatically in final layers.
Figure~\ref{fig:llama_explained_variance} reveals late-stage specialization for language identity.
Pairwise maps (Figures ~\ref{fig:llama_lang_map_en_cn}, ~\ref{fig:llama_lang_map_en_es}, ~\ref{fig:llama_lang_map_en_ru}, ~\ref{fig:llama_lang_map_en_hin}) confirm clean linear separability for all tested pairs.

\paragraph{Language prediction.} We train a logistic regression classifier on final-layer projections. Classifier trained only on the first principal component achieves near-perfect accuracy (0.96-1.00) across all language pairs and both models (Table~\ref{tab:clf}).
We observe decreased performance for the English-Spanish pair in both models, attributable to these languages being more closely related to each other.
\subsection{Language steering}

\paragraph{Next-token distribution evaluation.} We fit the steering module on Flores Plus with 200 parallel samples and test on 2,467 TED Talks \citep{banik2017tedtalks} samples with artificial code-switching.
Each sample is truncated to 20 words, split at the midpoint, and the second half is translated to the target language using Gemini 2.5 Flash via the OpenRouter API, producing texts that begin in English and switch mid-sentence.
Success is measured by comparing the next-token prediction distribution of the steered model against the English baseline via: KL divergence $\text{KL}(P_{\text{EN}} \| P_{\text{steered}})$, cosine similarity, and top-50 token overlap.

Table~\ref{tab:kl_cos_overlap} shows steering reduces KL divergence, increases cosine similarity and token overlap across all models and language pairs with a notable inefficiency on English-Hindi language pair.
The method shifts predictions from Spanish tokens (muy, bastante, tan) to their English equivalents (very, quite, so) while preserving semantic coherence (Table~\ref{tab:token_sample}).
We hypothesize that model-specific tokenization and training data balance play a key role. The Qwen model processed much more English and Chinese data, leading to more granular tokenization and better internal generalization for that pair.
At the same time, Hindi representation in the training corpus is usually much poorer.

\begin{table}[h]
  \centering
  \small
  \begin{tabular}{l l l}
    \hline
    Language pair & Qwen & Llama \\
    \hline
    English - Chinese & 10.15 / 0.02 / 3.69\% $\rightarrow$   & 9.05 / 0.05 / 4.47\% $\rightarrow$  \\
    & 4.52 / 0.22 / 26.46\% & 4.77 / 0.21 / 24.31\% \\
    English - Spanish & 7.25 / 0.12 / 11.41\% $\rightarrow$ & 6.52 / 0.12 / 11.45\% $\rightarrow$  \\
    & 4.90 / 0.18s / 20.23\% & 3.48 / 0.29 / 30.74\% \\
    English - Russian & 8.70 / 0.09 / 7.41\% $\rightarrow$  & 7.77 / 0.10 / 7.49\% $\rightarrow$  \\
    &5.40 / 0.23 / 25.45\% & 3.49 / 0.31 / 31.96\% \\
    English - Hindi & 9.60 / 0.04 / 11.23\% $\rightarrow$  & 8.06 / 0.06 / 6.54\% $\rightarrow$  \\
    &7.02 / 0.07 / 11.68\% & 6.13 / 0.16 / 25.88\% \\
    \hline
  \end{tabular}
  \caption{KL divergence / Cosine similarity / Top-50 token overlap (no steering $\rightarrow$ steering)}
  \label{tab:kl_cos_overlap}
\end{table}

\begin{table}[h]
  \centering
  \small
  \begin{tabular}{l c c c}
    \hline
    & & \multicolumn{2}{c}{Steered} \\
    \cmidrule(lr){3-4}
    Language pair & Unsteered & KL-opt. & CSI-opt. \\
    \hline
    English - Chinese  & 0.64 & 0.03 & 0.03 \\
    English - Spanish  & 0.62 & 0.23 & 0.11 \\
    English - Russian  & 0.59 & 0.02 & 0.01 \\
    English - Hindi    & 0.81 & 0.04 & 0.01 \\
    \hline
  \end{tabular}
  \caption{Code-Switching Index (CSI, $\downarrow$) of generated text (Llama-3.2-1B, $n=500$, code-switched input). English-only baseline: CSI~$= 0.02$. }
  \label{tab:gen_results}
\end{table}

\begin{table}
  \centering
  \small
  \begin{tabular}{l c c c}
    \hline
    & Qwen & \multicolumn{2}{c}{Llama} \\
    \cmidrule(lr){2-2} \cmidrule(lr){3-4}
    Language pair & KL-opt. & KL-opt. & CSI-opt. \\
    \hline
    English - Chinese & $-3.2$ & $-5.0$ & $-5.0$ \\
    English - Spanish & $-2.8$ & $-3.7$ & $-5.0$ \\
    English - Russian & $-2.2$ & $-3.5$ & $-4.0$ \\
    English - Hindi   & $+5.0$ & $-3.6$ & $-5.0$ \\
    \hline
  \end{tabular}
  \caption{Best steering strength coefficients. KL-optimized: from the distributional grid search; CSI-optimized: from the generation grid search.}
  \label{tab:steering_coeff}
\end{table}

\paragraph{Generation-based evaluation.} To complement the distributional analysis above, we evaluate steering on actual text generation using Llama-3.2-1B.
For each of 500 TED Talks samples per language pair, we generate 100-token continuations in three conditions: (A)~English-only input without steering, (B)~code-switched input without steering, and (C)~code-switched input with steering.
We measure code-switching in the generated output with the \emph{Code-Switching Index} (CSI)~\cite{gamback-das-2016-comparing} and a FastText classifier~\citep{joulin-etal-2017-bag}.

We evaluate two sets of steering coefficients (Table~\ref{tab:steering_coeff}).
Notably, the optimal steering coefficient (Table~\ref{tab:steering_coeff}) for Qwen English-Hindi is positive ($+5.0$), indicating the biggest variance comes not from the language component, but from some other component of unexplored nature.
Coefficients are highly fragile: small changes in magnitude can shift the model from fluent code-switched output to degenerate repetition.
Table~\ref{tab:qualitative} shows representative examples.

Table~\ref{tab:gen_results} shows that steering substantially reduces code-switching in generated text. CSI-optimized coefficients tend toward higher magnitudes, which improves language control but increases degeneration risk (see Limitations).

\section{Conclusion}

We introduced inference-time language steering in latent space.
By extracting language directions via PCA on parallel translations, our approach enables control over language identity (reduced KL divergence by up to 55\%, Code-Switching Index by 63--99\%) with negligible overhead - a single projection.
We also show that language identity concentrates in final layers with near-perfect linear separability (96-100\% classification accuracy).
Further work remains on the stability of steering coefficients (aggressive values can produce degenerate repetition) and generalization.

\section*{Limitations}

\begin{itemize}
  \item \textbf{Model scale and diversity.} Experiments limited to small models (1-1.5B parameters) and two architectures (Qwen2.5, Llama-3.2). Larger models and diverse architectures remain unexplored.

  \item \textbf{Incomplete reconstruction.} While the method achieves up to 55\% KL divergence reduction, the remaining gap indicates linear projections cannot fully reconstruct monolingual distributions, suggesting non-linear language-semantic interactions at the manifold level.

  \item \textbf{Dataset coverage.} Evaluation relies on Flores Plus and TED Talks (formal text). Conversational data, technical domains, and naturally occurring code-switching patterns are not tested. Our artificial code-switching setup produces mid-sentence switches at fixed positions, which may not reflect the morphosyntactic boundaries where natural code-switching typically occurs. Evaluation on naturalistic code-switching corpora such as the LinCE benchmark~\cite{aguilar-etal-2020-lince} is an important direction for future work.

  \item \textbf{Language typology.} Limited to Indo-European and Chinese languages. Low-resource languages, right-to-left scripts, and typologically distant languages (agglutinative, tonal) unexplored. Hindi showed the weakest improvement across both models, with KL reduction (24-27\%) but minimal gains in other distributional similarity metrics, suggesting tokenization or data balance issues.

  \item \textbf{Layer-wise steering.} Steering applied only to final layers based on empirical observation. Systematic exploration of multi-layer or adaptive steering strategies not conducted.

  \item \textbf{Evaluation metrics.} While we supplement distributional metrics (KL divergence) with a generation-based code-switching metric (CSI), we do not evaluate semantic preservation or downstream task performance. Human evaluation of output quality remains needed.

  \item \textbf{Fluency--control trade-off and coefficient fragility.} Generation-based evaluation reveals that aggressive steering coefficients, while effective at eliminating code-switching (CSI~$< 0.01$ for English-Russian and English-Hindi), can produce repetitive or degenerate output patterns. Steering coefficients are highly fragile. Small changes in magnitude shift behavior from fluent code-switched text to degenerate repetition. The CSI-optimized coefficients cluster near the maximum tested values ($-5.0$), suggesting that direct optimization of generation metrics pushes toward an aggressive regime. Combining steering with constrained decoding or adaptive coefficient scheduling could mitigate this issue.

  \item \textbf{Steering calibration.} Optimal coefficients require manual grid search per language pair and depend on the optimization target (distributional vs.\ generation-based). Principled or adaptive selection methods not developed.

  \item \textbf{Efficiency benchmarking.} No rigorous latency profiling across hardware configurations or batch sizes to validate overhead claims for production deployment.

  \item \textbf{Bilingual, English-only steering.} Our evaluation steers exclusively toward English in bilingual settings. Multilingual steering and bidirectional steering are natural extensions that we leave to future work. The multilingual PCA maps (Figures~\ref{fig:qwen_lang_map},~\ref{fig:llama_lang_map}) suggest the geometric structure supports such extensions, as multiple language clusters are simultaneously separable.
\end{itemize}

\section*{Acknowledgements}

The research was supported by the Russian Science Foundation grant No. 25-11-00355, https://rscf.ru/project/25-11-00355/.

\bibliography{custom}

@inproceedings{aguilar-etal-2020-lince,
  title     = {{L}in{CE}: A Centralized Benchmark for Linguistic Code-switching Evaluation},
  author    = {Aguilar, Gustavo  and
               Kar, Sudipta  and
               Solorio, Thamar},
  editor    = {Calzolari, Nicoletta  and
               B{\'e}chet, Fr{\'e}d{\'e}ric  and
               Blache, Philippe  and
               Choukri, Khalid  and
               Cieri, Christopher  and
               Declerck, Thierry  and
               Goggi, Sara  and
               Isahara, Hitoshi  and
               Maegaard, Bente  and
               Mariani, Joseph  and
               Mazo, H{\'e}l{\`e}ne  and
               Moreno, Asuncion  and
               Odijk, Jan  and
               Piperidis, Stelios},
  booktitle = {Proceedings of the Twelfth Language Resources and Evaluation Conference},
  month     = may,
  year      = {2020},
  address   = {Marseille, France},
  publisher = {European Language Resources Association},
  url       = {https://aclanthology.org/2020.lrec-1.223/},
  pages     = {1803--1813},
  language  = {eng},
  isbn      = {979-10-95546-34-4}
}

@inproceedings{bali-etal-2014-borrowing,
  title     = {``{I} am borrowing ya mixing ?'' An Analysis of {E}nglish-{H}indi Code Mixing in {F}acebook},
  author    = {Bali, Kalika  and
               Sharma, Jatin  and
               Choudhury, Monojit  and
               Vyas, Yogarshi},
  editor    = {Diab, Mona  and
               Hirschberg, Julia  and
               Fung, Pascale  and
               Solorio, Thamar},
  booktitle = {Proceedings of the First Workshop on Computational Approaches to Code Switching},
  month     = oct,
  year      = {2014},
  address   = {Doha, Qatar},
  publisher = {Association for Computational Linguistics},
  url       = {https://aclanthology.org/W14-3914/},
  doi       = {10.3115/v1/W14-3914},
  pages     = {116--126}
}

@misc{belrose2025leaceperfectlinearconcept,
  title         = {LEACE: Perfect linear concept erasure in closed form},
  author        = {Nora Belrose and David Schneider-Joseph and Shauli Ravfogel and Ryan Cotterell and Edward Raff and Stella Biderman},
  year          = {2025},
  eprint        = {2306.03819},
  archiveprefix = {arXiv},
  primaryclass  = {cs.LG},
  url           = {https://arxiv.org/abs/2306.03819}
}

@inproceedings{chi-etal-2020-finding,
  title     = {Finding Universal Grammatical Relations in Multilingual {BERT}},
  author    = {Chi, Ethan A.  and
               Hewitt, John  and
               Manning, Christopher D.},
  editor    = {Jurafsky, Dan  and
               Chai, Joyce  and
               Schluter, Natalie  and
               Tetreault, Joel},
  booktitle = {Proceedings of the 58th Annual Meeting of the {A}ssociation for {C}omputational {L}inguistics},
  month     = jul,
  year      = {2020},
  address   = {Online},
  publisher = {Association for Computational Linguistics},
  url       = {https://aclanthology.org/2020.acl-main.493/},
  doi       = {10.18653/v1/2020.acl-main.493},
  pages     = {5564--5577}
}

@inproceedings{conneau-etal-2020-emerging,
  title     = {Emerging Cross-lingual Structure in Pretrained Language Models},
  author    = {Conneau, Alexis  and
               Wu, Shijie  and
               Li, Haoran  and
               Zettlemoyer, Luke  and
               Stoyanov, Veselin},
  editor    = {Jurafsky, Dan  and
               Chai, Joyce  and
               Schluter, Natalie  and
               Tetreault, Joel},
  booktitle = {Proceedings of the 58th Annual Meeting of the {A}ssociation for {C}omputational {L}inguistics},
  month     = jul,
  year      = {2020},
  address   = {Online},
  publisher = {Association for Computational Linguistics},
  url       = {https://aclanthology.org/2020.acl-main.536/},
  doi       = {10.18653/v1/2020.acl-main.536},
  pages     = {6022--6034}
}

@misc{dathathri2020plugplaylanguagemodels,
  title         = {Plug and Play Language Models: A Simple Approach to Controlled Text Generation},
  author        = {Sumanth Dathathri and Andrea Madotto and Janice Lan and Jane Hung and Eric Frank and Piero Molino and Jason Yosinski and Rosanne Liu},
  year          = {2020},
  eprint        = {1912.02164},
  archiveprefix = {arXiv},
  primaryclass  = {cs.CL},
  url           = {https://arxiv.org/abs/1912.02164}
}

@inproceedings{gamback-das-2016-comparing,
  title     = {Comparing the Level of Code-Switching in Corpora},
  author    = {Gamb{\"a}ck, Bj{\"o}rn  and
               Das, Amitava},
  editor    = {Calzolari, Nicoletta  and
               Choukri, Khalid  and
               Declerck, Thierry  and
               Goggi, Sara  and
               Grobelnik, Marko  and
               Maegaard, Bente  and
               Mariani, Joseph  and
               Mazo, Helene  and
               Moreno, Asuncion  and
               Odijk, Jan  and
               Piperidis, Stelios},
  booktitle = {Proceedings of the Tenth International Conference on Language Resources and Evaluation ({LREC}'16)},
  month     = may,
  year      = {2016},
  address   = {Portoro{\v{z}}, Slovenia},
  publisher = {European Language Resources Association (ELRA)},
  url       = {https://aclanthology.org/L16-1292/},
  pages     = {1850--1855}
}

@misc{grattafiori2024llama3herdmodels,
  title         = {The Llama 3 Herd of Models},
  author        = {Aaron Grattafiori and Abhimanyu Dubey and Abhinav Jauhri and Abhinav Pandey and Abhishek Kadian and Ahmad Al-Dahle and Aiesha Letman and Akhil Mathur and Alan Schelten and Alex Vaughan and Amy Yang and Angela Fan and Anirudh Goyal and Anthony Hartshorn and Aobo Yang and Archi Mitra and Archie Sravankumar and Artem Korenev and Arthur Hinsvark and Arun Rao and Aston Zhang and Aurelien Rodriguez and Austen Gregerson and Ava Spataru and Baptiste Roziere and Bethany Biron and Binh Tang and Bobbie Chern and Charlotte Caucheteux and Chaya Nayak and Chloe Bi and Chris Marra and Chris McConnell and Christian Keller and Christophe Touret and Chunyang Wu and Corinne Wong and Cristian Canton Ferrer and Cyrus Nikolaidis and Damien Allonsius and Daniel Song and Danielle Pintz and Danny Livshits and Danny Wyatt and David Esiobu and Dhruv Choudhary and Dhruv Mahajan and Diego Garcia-Olano and Diego Perino and Dieuwke Hupkes and Egor Lakomkin and Ehab AlBadawy and Elina Lobanova and Emily Dinan and Eric Michael Smith and Filip Radenovic and Francisco Guzmán and Frank Zhang and Gabriel Synnaeve and Gabrielle Lee and Georgia Lewis Anderson and Govind Thattai and Graeme Nail and Gregoire Mialon and Guan Pang and Guillem Cucurell and Hailey Nguyen and Hannah Korevaar and Hu Xu and Hugo Touvron and Iliyan Zarov and Imanol Arrieta Ibarra and Isabel Kloumann and Ishan Misra and Ivan Evtimov and Jack Zhang and Jade Copet and Jaewon Lee and Jan Geffert and Jana Vranes and Jason Park and Jay Mahadeokar and Jeet Shah and Jelmer van der Linde and Jennifer Billock and Jenny Hong and Jenya Lee and Jeremy Fu and Jianfeng Chi and Jianyu Huang and Jiawen Liu and Jie Wang and Jiecao Yu and Joanna Bitton and Joe Spisak and Jongsoo Park and Joseph Rocca and Joshua Johnstun and Joshua Saxe and Junteng Jia and Kalyan Vasuden Alwala and Karthik Prasad and Kartikeya Upasani and Kate Plawiak and Ke Li and Kenneth Heafield and Kevin Stone and Khalid El-Arini and Krithika Iyer and Kshitiz Malik and Kuenley Chiu and Kunal Bhalla and Kushal Lakhotia and Lauren Rantala-Yeary and Laurens van der Maaten and Lawrence Chen and Liang Tan and Liz Jenkins and Louis Martin and Lovish Madaan and Lubo Malo and Lukas Blecher and Lukas Landzaat and Luke de Oliveira and Madeline Muzzi and Mahesh Pasupuleti and Mannat Singh and Manohar Paluri and Marcin Kardas and Maria Tsimpoukelli and Mathew Oldham and Mathieu Rita and Maya Pavlova and Melanie Kambadur and Mike Lewis and Min Si and Mitesh Kumar Singh and Mona Hassan and Naman Goyal and Narjes Torabi and Nikolay Bashlykov and Nikolay Bogoychev and Niladri Chatterji and Ning Zhang and Olivier Duchenne and Onur Çelebi and Patrick Alrassy and Pengchuan Zhang and Pengwei Li and Petar Vasic and Peter Weng and Prajjwal Bhargava and Pratik Dubal and Praveen Krishnan and Punit Singh Koura and Puxin Xu and Qing He and Qingxiao Dong and Ragavan Srinivasan and Raj Ganapathy and Ramon Calderer and Ricardo Silveira Cabral and Robert Stojnic and Roberta Raileanu and Rohan Maheswari and Rohit Girdhar and Rohit Patel and Romain Sauvestre and Ronnie Polidoro and Roshan Sumbaly and Ross Taylor and Ruan Silva and Rui Hou and Rui Wang and Saghar Hosseini and Sahana Chennabasappa and Sanjay Singh and Sean Bell and Seohyun Sonia Kim and Sergey Edunov and Shaoliang Nie and Sharan Narang and Sharath Raparthy and Sheng Shen and Shengye Wan and Shruti Bhosale and Shun Zhang and Simon Vandenhende and Soumya Batra and Spencer Whitman and Sten Sootla and Stephane Collot and Suchin Gururangan and Sydney Borodinsky and Tamar Herman and Tara Fowler and Tarek Sheasha and Thomas Georgiou and Thomas Scialom and Tobias Speckbacher and Todor Mihaylov and Tong Xiao and Ujjwal Karn and Vedanuj Goswami and Vibhor Gupta and Vignesh Ramanathan and Viktor Kerkez and Vincent Gonguet and Virginie Do and Vish Vogeti and Vítor Albiero and Vladan Petrovic and Weiwei Chu and Wenhan Xiong and Wenyin Fu and Whitney Meers and Xavier Martinet and Xiaodong Wang and Xiaofang Wang and Xiaoqing Ellen Tan and Xide Xia and Xinfeng Xie and Xuchao Jia and Xuewei Wang and Yaelle Goldschlag and Yashesh Gaur and Yasmine Babaei and Yi Wen and Yiwen Song and Yuchen Zhang and Yue Li and Yuning Mao and Zacharie Delpierre Coudert and Zheng Yan and Zhengxing Chen and Zoe Papakipos and Aaditya Singh and Aayushi Srivastava and Abha Jain and Adam Kelsey and Adam Shajnfeld and Adithya Gangidi and Adolfo Victoria and Ahuva Goldstand and Ajay Menon and Ajay Sharma and Alex Boesenberg and Alexei Baevski and Allie Feinstein and Amanda Kallet and Amit Sangani and Amos Teo and Anam Yunus and Andrei Lupu and Andres Alvarado and Andrew Caples and Andrew Gu and Andrew Ho and Andrew Poulton and Andrew Ryan and Ankit Ramchandani and Annie Dong and Annie Franco and Anuj Goyal and Aparajita Saraf and Arkabandhu Chowdhury and Ashley Gabriel and Ashwin Bharambe and Assaf Eisenman and Azadeh Yazdan and Beau James and Ben Maurer and Benjamin Leonhardi and Bernie Huang and Beth Loyd and Beto De Paola and Bhargavi Paranjape and Bing Liu and Bo Wu and Boyu Ni and Braden Hancock and Bram Wasti and Brandon Spence and Brani Stojkovic and Brian Gamido and Britt Montalvo and Carl Parker and Carly Burton and Catalina Mejia and Ce Liu and Changhan Wang and Changkyu Kim and Chao Zhou and Chester Hu and Ching-Hsiang Chu and Chris Cai and Chris Tindal and Christoph Feichtenhofer and Cynthia Gao and Damon Civin and Dana Beaty and Daniel Kreymer and Daniel Li and David Adkins and David Xu and Davide Testuggine and Delia David and Devi Parikh and Diana Liskovich and Didem Foss and Dingkang Wang and Duc Le and Dustin Holland and Edward Dowling and Eissa Jamil and Elaine Montgomery and Eleonora Presani and Emily Hahn and Emily Wood and Eric-Tuan Le and Erik Brinkman and Esteban Arcaute and Evan Dunbar and Evan Smothers and Fei Sun and Felix Kreuk and Feng Tian and Filippos Kokkinos and Firat Ozgenel and Francesco Caggioni and Frank Kanayet and Frank Seide and Gabriela Medina Florez and Gabriella Schwarz and Gada Badeer and Georgia Swee and Gil Halpern and Grant Herman and Grigory Sizov and Guangyi and Zhang and Guna Lakshminarayanan and Hakan Inan and Hamid Shojanazeri and Han Zou and Hannah Wang and Hanwen Zha and Haroun Habeeb and Harrison Rudolph and Helen Suk and Henry Aspegren and Hunter Goldman and Hongyuan Zhan and Ibrahim Damlaj and Igor Molybog and Igor Tufanov and Ilias Leontiadis and Irina-Elena Veliche and Itai Gat and Jake Weissman and James Geboski and James Kohli and Janice Lam and Japhet Asher and Jean-Baptiste Gaya and Jeff Marcus and Jeff Tang and Jennifer Chan and Jenny Zhen and Jeremy Reizenstein and Jeremy Teboul and Jessica Zhong and Jian Jin and Jingyi Yang and Joe Cummings and Jon Carvill and Jon Shepard and Jonathan McPhie and Jonathan Torres and Josh Ginsburg and Junjie Wang and Kai Wu and Kam Hou U and Karan Saxena and Kartikay Khandelwal and Katayoun Zand and Kathy Matosich and Kaushik Veeraraghavan and Kelly Michelena and Keqian Li and Kiran Jagadeesh and Kun Huang and Kunal Chawla and Kyle Huang and Lailin Chen and Lakshya Garg and Lavender A and Leandro Silva and Lee Bell and Lei Zhang and Liangpeng Guo and Licheng Yu and Liron Moshkovich and Luca Wehrstedt and Madian Khabsa and Manav Avalani and Manish Bhatt and Martynas Mankus and Matan Hasson and Matthew Lennie and Matthias Reso and Maxim Groshev and Maxim Naumov and Maya Lathi and Meghan Keneally and Miao Liu and Michael L. Seltzer and Michal Valko and Michelle Restrepo and Mihir Patel and Mik Vyatskov and Mikayel Samvelyan and Mike Clark and Mike Macey and Mike Wang and Miquel Jubert Hermoso and Mo Metanat and Mohammad Rastegari and Munish Bansal and Nandhini Santhanam and Natascha Parks and Natasha White and Navyata Bawa and Nayan Singhal and Nick Egebo and Nicolas Usunier and Nikhil Mehta and Nikolay Pavlovich Laptev and Ning Dong and Norman Cheng and Oleg Chernoguz and Olivia Hart and Omkar Salpekar and Ozlem Kalinli and Parkin Kent and Parth Parekh and Paul Saab and Pavan Balaji and Pedro Rittner and Philip Bontrager and Pierre Roux and Piotr Dollar and Polina Zvyagina and Prashant Ratanchandani and Pritish Yuvraj and Qian Liang and Rachad Alao and Rachel Rodriguez and Rafi Ayub and Raghotham Murthy and Raghu Nayani and Rahul Mitra and Rangaprabhu Parthasarathy and Raymond Li and Rebekkah Hogan and Robin Battey and Rocky Wang and Russ Howes and Ruty Rinott and Sachin Mehta and Sachin Siby and Sai Jayesh Bondu and Samyak Datta and Sara Chugh and Sara Hunt and Sargun Dhillon and Sasha Sidorov and Satadru Pan and Saurabh Mahajan and Saurabh Verma and Seiji Yamamoto and Sharadh Ramaswamy and Shaun Lindsay and Shaun Lindsay and Sheng Feng and Shenghao Lin and Shengxin Cindy Zha and Shishir Patil and Shiva Shankar and Shuqiang Zhang and Shuqiang Zhang and Sinong Wang and Sneha Agarwal and Soji Sajuyigbe and Soumith Chintala and Stephanie Max and Stephen Chen and Steve Kehoe and Steve Satterfield and Sudarshan Govindaprasad and Sumit Gupta and Summer Deng and Sungmin Cho and Sunny Virk and Suraj Subramanian and Sy Choudhury and Sydney Goldman and Tal Remez and Tamar Glaser and Tamara Best and Thilo Koehler and Thomas Robinson and Tianhe Li and Tianjun Zhang and Tim Matthews and Timothy Chou and Tzook Shaked and Varun Vontimitta and Victoria Ajayi and Victoria Montanez and Vijai Mohan and Vinay Satish Kumar and Vishal Mangla and Vlad Ionescu and Vlad Poenaru and Vlad Tiberiu Mihailescu and Vladimir Ivanov and Wei Li and Wenchen Wang and Wenwen Jiang and Wes Bouaziz and Will Constable and Xiaocheng Tang and Xiaojian Wu and Xiaolan Wang and Xilun Wu and Xinbo Gao and Yaniv Kleinman and Yanjun Chen and Ye Hu and Ye Jia and Ye Qi and Yenda Li and Yilin Zhang and Ying Zhang and Yossi Adi and Youngjin Nam and Yu and Wang and Yu Zhao and Yuchen Hao and Yundi Qian and Yunlu Li and Yuzi He and Zach Rait and Zachary DeVito and Zef Rosnbrick and Zhaoduo Wen and Zhenyu Yang and Zhiwei Zhao and Zhiyu Ma},
  year          = {2024},
  eprint        = {2407.21783},
  archiveprefix = {arXiv},
  primaryclass  = {cs.AI},
  url           = {https://arxiv.org/abs/2407.21783}
}

@misc{li2024inferencetimeinterventionelicitingtruthful,
  title         = {Inference-Time Intervention: Eliciting Truthful Answers from a Language Model},
  author        = {Kenneth Li and Oam Patel and Fernanda Viégas and Hanspeter Pfister and Martin Wattenberg},
  year          = {2024},
  eprint        = {2306.03341},
  archiveprefix = {arXiv},
  primaryclass  = {cs.LG},
  url           = {https://arxiv.org/abs/2306.03341}
}

@misc{liu2021dexpertsdecodingtimecontrolledtext,
  title         = {DExperts: Decoding-Time Controlled Text Generation with Experts and Anti-Experts},
  author        = {Alisa Liu and Maarten Sap and Ximing Lu and Swabha Swayamdipta and Chandra Bhagavatula and Noah A. Smith and Yejin Choi},
  year          = {2021},
  eprint        = {2105.03023},
  archiveprefix = {arXiv},
  primaryclass  = {cs.CL},
  url           = {https://arxiv.org/abs/2105.03023}
}

@article{scaling-2024,
  author        = {Costa-juss{\`a}, Marta R. and Cross, James and {\c C}elebi, Onur and Elbayad, Maha and Heafield, Kenneth and Heffernan, Kevin and Kalbassi, Elahe and Lam, Janice and Licht, Daniel and Maillard, Jean and Sun, Anna and Wang, Skyler and Wenzek, Guillaume and Youngblood, Al and Akula, Bapi and Barrault, Loic and Gonzalez, Gabriel Mejia and Hansanti, Prangthip and Hoffman, John and Jarrett, Semarley and Sadagopan, Kaushik Ram and Rowe, Dirk and Spruit, Shannon and Tran, Chau and Andrews, Pierre and Ayan, Necip Fazil and Bhosale, Shruti and Edunov, Sergey and Fan, Angela and Gao, Cynthia and Goswami, Vedanuj and Guzm{\'a}n, Francisco and Koehn, Philipp and Mourachko, Alexandre and Ropers, Christophe and Saleem, Safiyyah and Schwenk, Holger and Wang, Jeff and NLLB Team},
  date          = {2024/06/01},
  doi           = {10.1038/s41586-024-07335-x},
  id            = {Costa-juss{\`a}2024},
  isbn          = {1476-4687},
  journal       = {Nature},
  number        = {8018},
  pages         = {841--846},
  title         = {Scaling neural machine translation to 200 languages},
  url           = {https://doi.org/10.1038/s41586-024-07335-x},
  volume        = {630},
  year          = {2024}
}

@misc{qwen2025qwen25technicalreport,
  title         = {Qwen2.5 Technical Report},
  author        = {Qwen and : and An Yang and Baosong Yang and Beichen Zhang and Binyuan Hui and Bo Zheng and Bowen Yu and Chengyuan Li and Dayiheng Liu and Fei Huang and Haoran Wei and Huan Lin and Jian Yang and Jianhong Tu and Jianwei Zhang and Jianxin Yang and Jiaxi Yang and Jingren Zhou and Junyang Lin and Kai Dang and Keming Lu and Keqin Bao and Kexin Yang and Le Yu and Mei Li and Mingfeng Xue and Pei Zhang and Qin Zhu and Rui Men and Runji Lin and Tianhao Li and Tianyi Tang and Tingyu Xia and Xingzhang Ren and Xuancheng Ren and Yang Fan and Yang Su and Yichang Zhang and Yu Wan and Yuqiong Liu and Zeyu Cui and Zhenru Zhang and Zihan Qiu},
  year          = {2025},
  eprint        = {2412.15115},
  archiveprefix = {arXiv},
  primaryclass  = {cs.CL},
  url           = {https://arxiv.org/abs/2412.15115}
}

@inproceedings{ravfogel-etal-2020-null,
  title     = {Null It Out: Guarding Protected Attributes by Iterative Nullspace Projection},
  author    = {Ravfogel, Shauli  and
               Elazar, Yanai  and
               Gonen, Hila  and
               Twiton, Michael  and
               Goldberg, Yoav},
  editor    = {Jurafsky, Dan  and
               Chai, Joyce  and
               Schluter, Natalie  and
               Tetreault, Joel},
  booktitle = {Proceedings of the 58th Annual Meeting of the {A}ssociation for {C}omputational {L}inguistics},
  month     = jul,
  year      = {2020},
  address   = {Online},
  publisher = {Association for Computational Linguistics},
  url       = {https://aclanthology.org/2020.acl-main.647/},
  doi       = {10.18653/v1/2020.acl-main.647},
  pages     = {7237--7256}
}

@inproceedings{ryan-etal-2024-unintended,
  title     = {Unintended Impacts of {LLM} Alignment on Global Representation},
  author    = {Ryan, Michael J  and
               Held, William  and
               Yang, Diyi},
  editor    = {Ku, Lun-Wei  and
               Martins, Andre  and
               Srikumar, Vivek},
  booktitle = {Proceedings of the 62nd Annual Meeting of the {A}ssociation for {C}omputational {L}inguistics (Volume 1: Long Papers)},
  month     = aug,
  year      = {2024},
  address   = {Bangkok, Thailand},
  publisher = {Association for Computational Linguistics},
  url       = {https://aclanthology.org/2024.acl-long.853/},
  doi       = {10.18653/v1/2024.acl-long.853},
  pages     = {16121--16140}
}

@misc{turner2024steeringlanguagemodelsactivation,
  title         = {Steering Language Models With Activation Engineering},
  author        = {Alexander Matt Turner and Lisa Thiergart and Gavin Leech and David Udell and Juan J. Vazquez and Ulisse Mini and Monte MacDiarmid},
  year          = {2024},
  eprint        = {2308.10248},
  archiveprefix = {arXiv},
  primaryclass  = {cs.CL},
  url           = {https://arxiv.org/abs/2308.10248}
}

@inproceedings{wendler-etal-2024-llamas,
  title     = {Do Llamas Work in {E}nglish? On the Latent Language of Multilingual Transformers},
  author    = {Wendler, Chris  and
               Veselovsky, Veniamin  and
               Monea, Giovanni  and
               West, Robert},
  editor    = {Ku, Lun-Wei  and
               Martins, Andre  and
               Srikumar, Vivek},
  booktitle = {Proceedings of the 62nd Annual Meeting of the {A}ssociation for {C}omputational {L}inguistics (Volume 1: Long Papers)},
  month     = aug,
  year      = {2024},
  address   = {Bangkok, Thailand},
  publisher = {Association for Computational Linguistics},
  url       = {https://aclanthology.org/2024.acl-long.820/},
  doi       = {10.18653/v1/2024.acl-long.820},
  pages     = {15366--15394}
}

@inproceedings{yang-etal-2021-simple,
  title     = {A Simple and Effective Method To Eliminate the Self Language Bias in Multilingual Representations},
  author    = {Yang, Ziyi  and
               Yang, Yinfei  and
               Cer, Daniel  and
               Darve, Eric},
  editor    = {Moens, Marie-Francine  and
               Huang, Xuanjing  and
               Specia, Lucia  and
               Yih, Scott Wen-tau},
  booktitle = {Proceedings of the 2021 Conference on Empirical Methods in Natural Language Processing},
  month     = nov,
  year      = {2021},
  address   = {Online and Punta Cana, Dominican Republic},
  publisher = {Association for Computational Linguistics},
  url       = {https://aclanthology.org/2021.emnlp-main.470/},
  doi       = {10.18653/v1/2021.emnlp-main.470},
  pages     = {5825--5832}
}

@inproceedings{yang-klein-2021-fudge,
  title     = {{FUDGE}: Controlled Text Generation With Future Discriminators},
  author    = {Yang, Kevin  and
               Klein, Dan},
  editor    = {Toutanova, Kristina  and
               Rumshisky, Anna  and
               Zettlemoyer, Luke  and
               Hakkani-Tur, Dilek  and
               Beltagy, Iz  and
               Bethard, Steven  and
               Cotterell, Ryan  and
               Chakraborty, Tanmoy  and
               Zhou, Yichao},
  booktitle = {Proceedings of the 2021 Conference of the {N}orth {A}merican Chapter of the {A}ssociation for {C}omputational {L}inguistics: Human Language Technologies},
  month     = jun,
  year      = {2021},
  address   = {Online},
  publisher = {Association for Computational Linguistics},
  url       = {https://aclanthology.org/2021.naacl-main.276/},
  doi       = {10.18653/v1/2021.naacl-main.276},
  pages     = {3511--3535}
}

@misc{yoo2025codeswitchingredteamingllmevaluation,
  title         = {Code-Switching Red-Teaming: LLM Evaluation for Safety and Multilingual Understanding},
  author        = {Haneul Yoo and Yongjin Yang and Hwaran Lee},
  year          = {2025},
  eprint        = {2406.15481},
  archiveprefix = {arXiv},
  primaryclass  = {cs.AI},
  url           = {https://arxiv.org/abs/2406.15481}
}

@misc{zou2025representationengineeringtopdownapproach,
  title         = {Representation Engineering: A Top-Down Approach to AI Transparency},
  author        = {Andy Zou and Long Phan and Sarah Chen and James Campbell and Phillip Guo and Richard Ren and Alexander Pan and Xuwang Yin and Mantas Mazeika and Ann-Kathrin Dombrowski and Shashwat Goel and Nathaniel Li and Michael J. Byun and Zifan Wang and Alex Mallen and Steven Basart and Sanmi Koyejo and Dawn Song and Matt Fredrikson and J. Zico Kolter and Dan Hendrycks},
  year          = {2025},
  eprint        = {2310.01405},
  archiveprefix = {arXiv},
  primaryclass  = {cs.LG},
  url           = {https://arxiv.org/abs/2310.01405}
}

@inproceedings{marchisio-etal-2024-understanding,
  title     = {Understanding and Mitigating Language Confusion in {LLM}s},
  author    = {Marchisio, Kelly  and
               Ko, Wei-Yin  and
               Berard, Alexandre  and
               Dehaze, Th{\'e}o  and
               Ruder, Sebastian},
  editor    = {Al-Onaizan, Yaser  and
               Bansal, Mohit  and
               Chen, Yun-Nung},
  booktitle = {Proceedings of the 2024 Conference on Empirical Methods in Natural Language Processing},
  month     = nov,
  year      = {2024},
  address   = {Miami, Florida, USA},
  publisher = {Association for Computational Linguistics},
  url       = {https://aclanthology.org/2024.emnlp-main.380/},
  doi       = {10.18653/v1/2024.emnlp-main.380},
  pages     = {6653--6677}
}

@misc{zhao2024largelanguagemodelshandle,
  title         = {How do Large Language Models Handle Multilingualism?},
  author        = {Yiran Zhao and Wenxuan Zhang and Guizhen Chen and Kenji Kawaguchi and Lidong Bing},
  year          = {2024},
  eprint        = {2402.18815},
  archiveprefix = {arXiv},
  primaryclass  = {cs.CL},
  url           = {https://arxiv.org/abs/2402.18815}
}

@misc{banik2017tedtalks,
  title        = {{TED} Talks},
  author       = {Banik, Rounak},
  year         = {2017},
  howpublished = {Kaggle dataset},
  url          = {https://www.kaggle.com/datasets/rounakbanik/ted-talks}
}
\bibliographystyle{plainnat}


\appendix
\section{Token overlap example}

\begin{table}[h]
  \small
  \centering
  \begin{tabular}{l l l}
    \hline
    Original & Unsteered & Steered \\
    \hline
    still & muy & very \\
    going & un  & surprised \\
    not & bastante & pretty \\
    so & tan & starting \\
    a & sor & working \\
    just & más & content \\
    really & a & rather \\
    in & empez & impressed \\
    very & de & quite \\
    sitting & content & -- \\
    quite & en & great \\
    pretty & enc & beginning \\
    here & aquí & answering \\
    thinking & pens & almost\\
    looking & realmente & fully \\
    kind & dese & \_\_\_\_ \\
    getting & completamente & \_\_ \\
    having & feliz & even \\
    actually & org & heart \\
    surprised & seguro & trying \\
    \hline
  \end{tabular}
  \caption{Top 20 token sample (Qwen, English-Spanish)}
  \label{tab:token_sample}
\end{table}

\section{Language classification results}

\begin{table}[h]
  \centering
  \begin{tabular}{l l l}
    \hline
    Language pair & Qwen & Llama \\
    \hline
    English - Chinese & 0.98 & 0.98 \\
    English - Spanish & 0.96 & 0.96 \\
    English - Russian & 0.99 & 0.99 \\
    English - Hindi & 0.98 & 1.00 \\
    \hline
  \end{tabular}
  \caption{Last layer language classifier accuracy}
  \label{tab:clf}
\end{table}

\section{Steering fragility example}

\begin{table}[ht]
  \centering
  \small
  \begin{tabular}{p{0.14\linewidth} p{0.80\linewidth}}
    \hline
    \multicolumn{2}{l}{\textbf{en-es:} \textit{``...I took on the task to ense\~{n}ar desarrollo global...''}} \\
    \hline
    Unsteered & \textit{de que un grupo de estudiantes suecos, que hab\'{i}an sido enviados a la Universidad de Harvard, me escribieron para pedir ayuda...} \\
    Steered & \textit{de that I had the opportunity to work with the first two editions of the course and I was very inspired by it...} \\
    \hline
    \multicolumn{2}{l}{\textbf{en-ru:} \textit{``...About 10 years ago, I took on the task to} [Russian text]\textit{...''}} \\
    \hline
    Degenerate steered & \textit{first year of the first year of the first year of the first year of the first year...} \\
    \hline
  \end{tabular}
  \caption{Generated continuations (Llama-3.2-1B, KL-optimized). Top: steering successfully redirects Spanish to coherent English. Bottom: steering eliminates Russian but produces degenerate repetition.}
  \label{tab:qualitative}
\end{table}

\end{document}